\documentclass[runningheads]{llncs}
\usepackage[T1]{fontenc}
\usepackage{graphicx}
\usepackage{bbding}
\usepackage{amsmath}
\usepackage{tabularx}
\graphicspath{{img}}
\begin{document}
\title{LGI-DETR: Local-Global Interaction for UAV Object Detection}
%
%
\author{Zifa Chen}
\authorrunning{Z. Chen et al.}
%
\institute{School of Computer Science and Technology, Hangzhou Dianzi University, China 
 \email{zfchen@hdu.edu.cn} 
}
\maketitle              
\begin{abstract}
        UAV has been widely used in various fields. However, most of the existing object detectors used in drones are not end-to-end and require the design of various complex components and careful fine-tuning. 
        Most of the existing end-to-end object detectors are designed for natural scenes.
        It is not ideal to apply them directly to UAV images. In order to solve the above challenges, we design an local-global information interaction DETR for UAVs, namely LGI-DETR. Cross-layer bidirectional low-level and high-level feature information enhancement, this fusion method is effective especially in the field of small objection detection. 
        At the initial stage of encoder, we propose a local spatial enhancement module (LSE), which enhances the low-level rich local spatial information into the high-level feature, and reduces the loss of local information in the transmission process of high-level information. 
        At the final stage of the encoder, we propose a novel global information injection module (GII) designed to integrate rich high-level global semantic representations with low-level feature maps.
        This hierarchical fusion mechanism effectively addresses the inherent limitations of local receptive fields by propagating contextual information across the feature hierarchy. 
        Experimental results on two challenging UAV image object detection benchmarks, VisDrone2019 and UAVDT, show that our proposed model outperforms the SOTA model. Compared to the baseline model, AP and AP$_{50}$ improved by 1.9\% and 2.4\%, respectively.
\end{abstract}

    \begin{figure}[t]
        \centering
        \includegraphics[width=\linewidth]{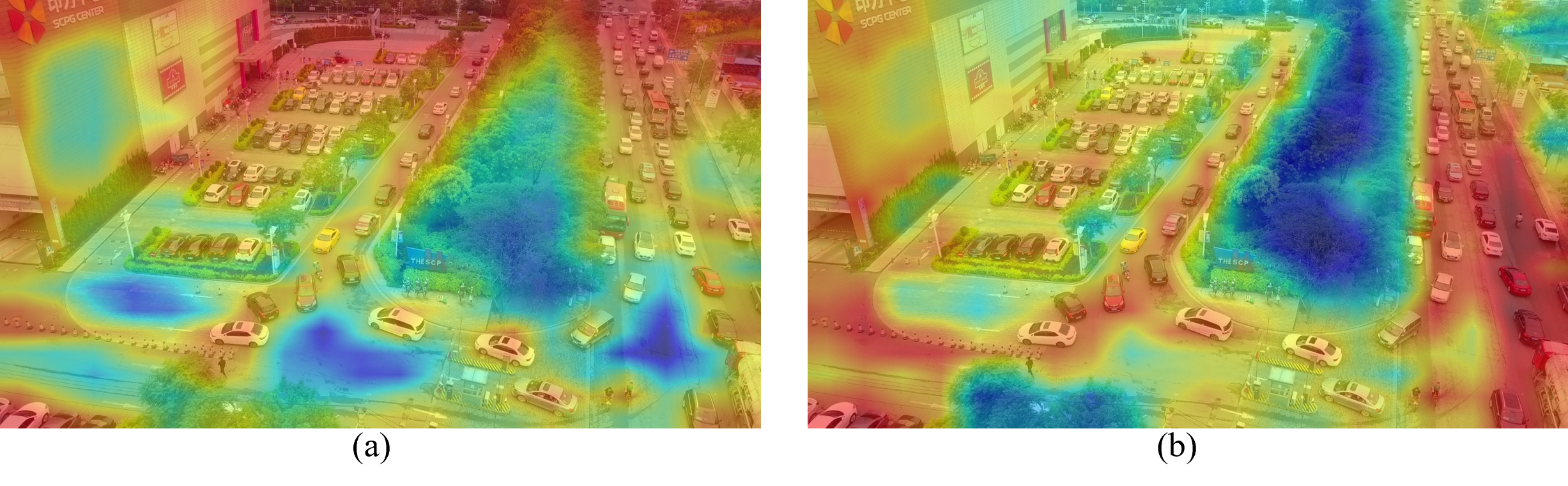}
        \caption{(a) The heatmap of baseline. (b) The heatmap of LGI-DETR. The
        brighter areas in the heatmap indicate stronger attention by the model. Our
        model shows more attention on objects than baseline model.}
        \label{fig:heatmap}
    \end{figure}

\section{Introduction}

Object detection is a fundamental task in computer vision, aiming to localize and identify specific objects within images or videos. It has been widely applied in autonomous driving, security surveillance, scene understanding, augmented reality, and video retrieval~\cite{ijaz2023uav}. Recent advances in detection algorithms, such as Faster R-CNN~\cite{ren2016faster}, YOLO series~\cite{wang2024yolov10}, and DETR~\cite{carion2020end}, have significantly improved detection accuracy and efficiency. 

However, Unmanned Aerial Vehicle (UAV) image-based object detection faces unique challenges due to the imaging perspective and complex scenes~\cite{lai2023real}. UAV images often contain dense and small objects, many occupying less than $32\times32$ pixels~\cite{tong2020recent}, resulting in limited semantic information and low feature resolution. Additionally, objects exhibit large scale variations, spatial heterogeneity, occlusions, and are susceptible to distortion and environmental noise. These factors severely degrade detection performance and demand feature extractors capable of capturing both local spatial details and global semantic context.

To address these challenges, various methods have been proposed, including advanced feature extraction~\cite{cui2023skip}, complex feature pyramid networks (FPN)~\cite{guo2023save}, multi-head detection~\cite{zhu2021tph}, customized attention mechanisms~\cite{leng2022pareto}, and two-stage pipelines~\cite{fang2023enhancing}. Region-of-interest-based cropping strategies like ESOD~\cite{liu2024esod} and YOLC~\cite{liu2024yolc} also attempt to suppress background interference. However, these methods heavily rely on handcrafted designs and require extensive hyperparameter tuning, such as anchor boxes and Non-Maximum Suppression (NMS)~\cite{tong2020recent}, which hampers scalability and deployment efficiency.

End-to-end detectors represented by DETR~\cite{carion2020end} reformulate object detection as a set prediction problem, eliminating the need for NMS. Deformable DETR~\cite{zhu2020deformable} alleviates DETR's inefficiency in small object detection, while RT-DETR~\cite{zhao2024detrs} further balances accuracy and speed. Nonetheless, these models still struggle with UAV-specific challenges due to limited adaptation to small objects and complex aerial scenes.
Inspired by recent advances in multi-modal conditional models and diffusion-based generative framework, which demonstrate strong abilities in learning rich semantic representations and handling complex scenes, we propose LGI-DETR. Our model enhances RT-DETR by integrating a Local Spatial Enhancement (LSE) module and a Global Information Injection (GII) module. The LSE module refines low-level spatial features, while GII injects high-level semantic context, enabling effective local-global feature interaction.

Our contributions are summarized as follows:
\begin{itemize}
    \item[$\bullet$] We propose LGI-DETR, an efficient end-to-end UAV object detection model that enhances local-global feature interaction without increasing computational complexity.
    \item[$\bullet$] The Local Spatial Enhancement module strengthens spatial details in early encoder stages, preserving fine-grained object features.
    \item[$\bullet$] The Global Information Injection module enriches semantic understanding at later stages, improving detection performance on small and occluded objects.
    \item[$\bullet$] Experiments on VisDrone2019 and UAVDT datasets demonstrate that LGI-DETR achieves superior performance over RT-DETR-r18, especially on AP and AP$_{50}$ metrics.
\end{itemize}

\section{Related Work}\label{rw}

\subsection{Small Object Detection in Aerial Imagery}
Small object detection in UAV imagery is inherently challenging due to limited pixel coverage, scale variations, and complex backgrounds. Recent works adopt coarse-to-fine strategies and region selection mechanisms~\cite{yang2019clustered,liu2024esod,liu2024yolc} to improve detection accuracy. However, these approaches often increase computational overhead and sacrifice real-time performance.
SGMFNet~\cite{zhang2023self} introduces global-local feature guidance and reverse residual enhancement, effectively extracting small object features. CEASC~\cite{du2023adaptive} employs sparse convolutions and adaptive masking to optimize resource-constrained UAV platforms. Despite these improvements, conventional methods rely heavily on NMS, adding latency and computation.
End-to-end models like DETR~\cite{carion2020end} eliminate NMS but struggle with small objects. Recent advancements in aerial object detection, such as LR-FPN~\cite{li2024lr} and Selective Frequency Interaction Networks (SFI-Net)~\cite{weng2024enhancing}, demonstrate the importance of hierarchical feature enhancement and frequency-based attention for robust UAV detection. Similarly, conditional generative models~\cite{shen2025long,shen2024imagpose} have shown great potential in handling complex visual scenarios, inspiring our local-global enhancement design.

\subsection{Feature Fusion and Multi-scale Representation}
Effective feature fusion is critical for balancing spatial precision and semantic richness. Methods like RepBi-PAN~\cite{li2023yolov6} and Gold-YOLO~\cite{wang2023gold} enhance feature aggregation through bidirectional concatenation and gather-distribute mechanisms. AFPN~\cite{yang2023afpn} enables non-adjacent layer interactions, bridging the semantic gap across scales.
However, most existing methods overlook the alignment between spatial and semantic features, leading to suboptimal performance in UAV scenarios. RT-DETR~\cite{zhao2024detrs} introduces a hybrid encoder but still lacks effective local feature enhancement.
Inspired by advancements in virtual dressing~\cite{shen2024imagdressing} and story visualization~\cite{shen2024boosting}, where multi-scale conditioning improves generation quality, our LGI-DETR incorporates LSE and GII modules. These modules establish complementary local-global relationships, enhancing the model's capability to process UAV imagery with dense small objects and complex spatial distributions.

\section{Proposed Method}
    This section introduces the proposed LGI-DETR model. The model is designed to improve the detection performance of UAV images. 
    The model is based on RT-DETR, which is an efficient transformer-based end-to-end detector. We have carefully designed the encoder part. 
    In the following sections, we provide a detailed description of the proposed model. 
    First, \ref{met:overview} introduce the overall structure of the model, then Section \ref{met:LSE} and Section \ref{met:GII}, describe in detail the two novel modules of the model: the Local Spatial Enhancement module (LSE) and the Global Information Injection module (GII). 

    \begin{figure}[t]
        \centering
        \includegraphics[width=1\linewidth]{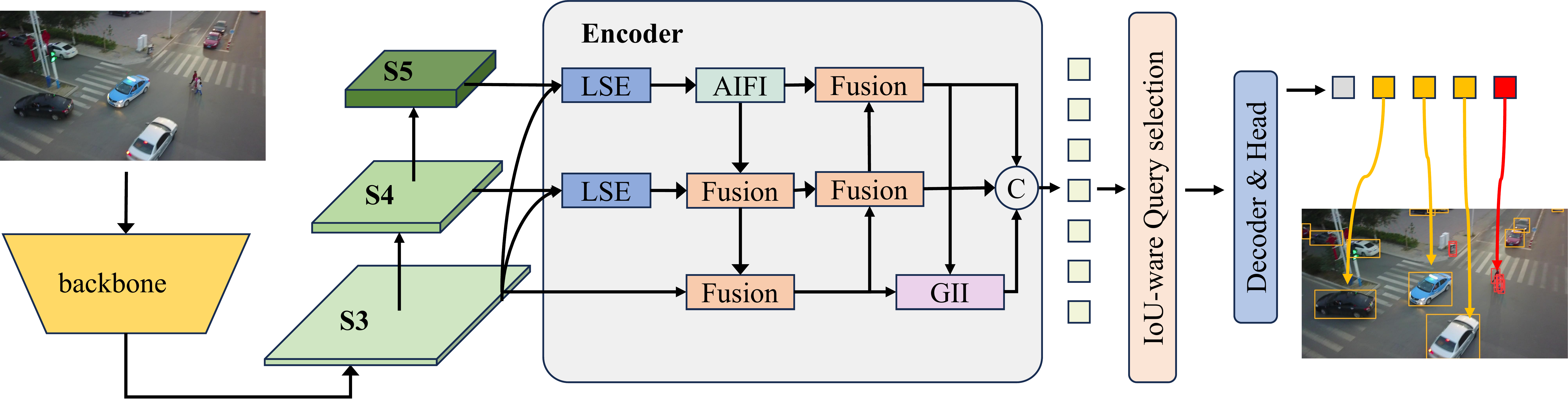}
        \caption{Overview of the LGI-DETR. Firstly, the multi-scale feature of
        the image is extracted by backbone. Then the multi-scale feature is input
        into the encoder for feature representation. Finally, the detection head
        outputs the object detection results for the generated object query. LSE
        represent the Local Spatial Enhancement module; GII represent the Global
        Information Injection module; AIFI represent the Attentionbased Intra-scale
        Feature Interaction. }
        \label{fig:lgi-detr}
    \end{figure}

    \subsection{Overview Architecture}
    \label{met:overview} Fig.~\ref{fig:lgi-detr} shows the overall structure of the
    proposed LGI-DETR, which comprises three main components: the backbone,
    efficient encoder, and detection head. The backbone is responsible for
    extracting the features of the input image. The efficient encoder is designed
    to fushion the features output by the backbone. And the detection head is
    responsible for locating and classifying the objects in the image. We enhance
    the model with two key modules: the Local Spatial Enhancement module (LSE)
    and the Global Information Injection module (GII). The LSE module enhances the
    local spatial information of the low-level features and fuses them into the high-level
    features. The GII module enhances the global semantic information of the
    high-level features and fuses them into the low-level features. The proposed
    LGI-DETR model is designed to improve the detection performance of UAV images.

    \subsection{Local Spatial Enhancement}
    \label{met:LSE} RT-DETR uses the self-attention mechanism to capture the
    correlation between the features output at the highest level of the backbone,
    so that the model can detect and classify the object in the image more accurately.
    RT-DETR believes that there is no need to interact with multi-scale features before self-attention,
    because there may be repetitive and confusing interactions. With the
    increase of depth and convolution operation, backbone leads to the loss of local spatial information in features. 
    The inherent loss of local spatial information in high-level feature representations significantly impairs the self-attention mechanism's capacity to effectively capture spatial correlation during feature interaction, consequently constraining the model's detection performance for small objects.
    Local spatial information is important for the detection of small objects.  Foreground objects typically occupy minimal spatial regions, while high-level features process foreground and background regions with equal emphasis, thereby significantly amplifying background interference in the feature representation.
    Such conditions critically impair the self-attention mechanism's ability to establish robust feature correlations, thereby limiting the model's efficacy in accurately localizing and classifying small objects.

    \begin{figure}[t]
        \centering
        \includegraphics[width=0.8\linewidth]{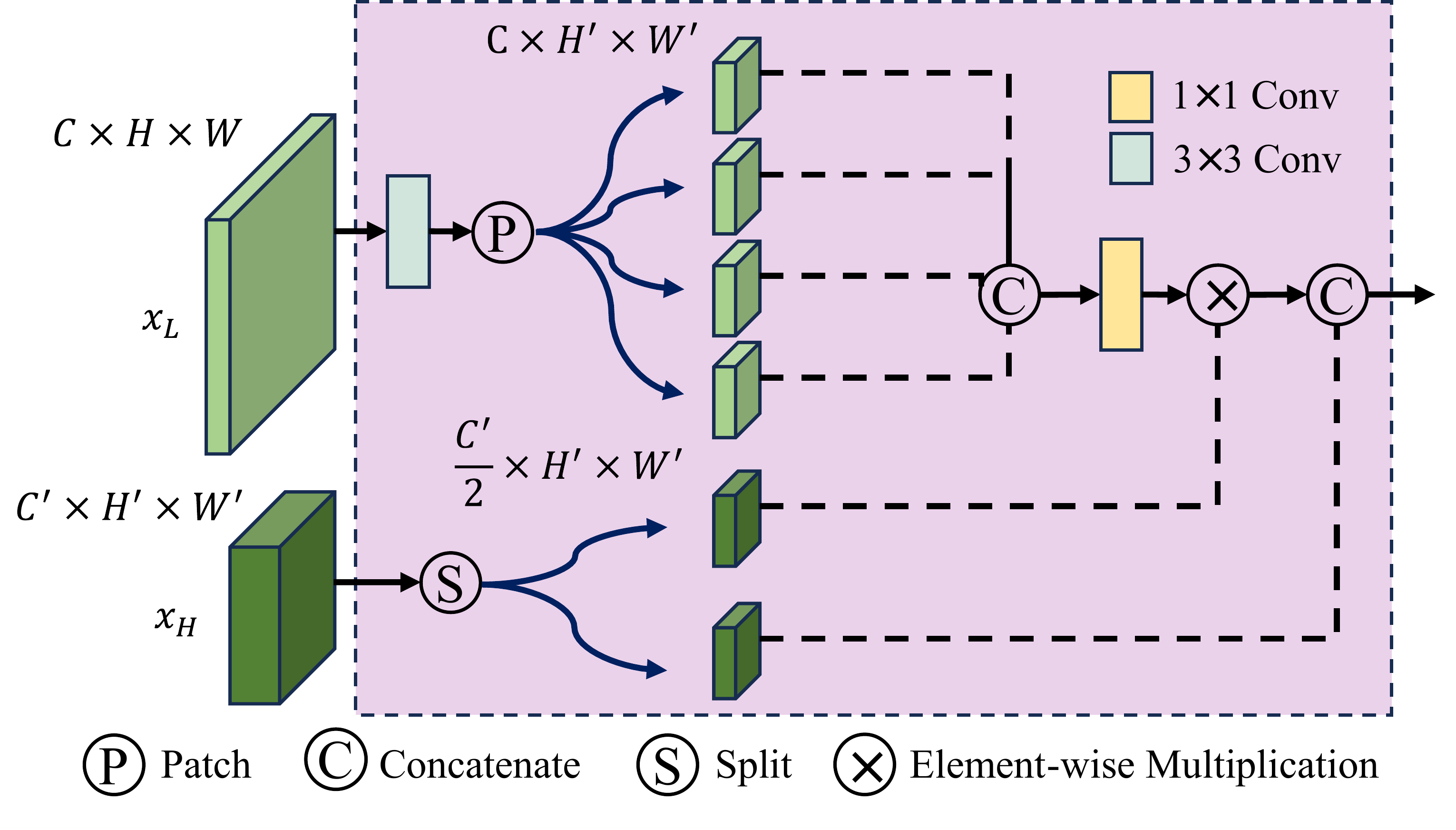}
        \caption{ Overview of LSE, which comprises two branches. The first branch
        is the low-level feature weight extraction operation, and the second
        branch is to fuse the extracted low-level feature weight into high-level
        features. The low-level feature weights derived through the patch merge operation are subsequently integrated with selective high-level channel features.}
        \label{fig:lse}
    \end{figure}

    However, when dealing with complex situations and small object tasks, low-level local spatial information and details are crucial to improve the accuracy of the model.
    High-level features are usually coarse-grained and rich in semantic information, while low-level features are fine-grained and retain more local spatial details.
    In order to solve the problem that the input encoder features lack low-level
    local spatial information, we propose a local spatial information enhancement (LSE) module before the encoder, as shown in Fig.~\ref{fig:lse}. 
    Before the top-down feature fusion, we introduce LSE to make the fused information contain more local spatial information of small objects. Specifically, as shown in Fig.~\ref{fig:lgi-detr}, before the encoder part, the bottom feature S3 containing more local spatial information is first enhanced and merged into the high-level S4 and S5 to obtain enhanced S4' and S5'. The encoder part further encodes the features of S4' and S5'.
    This makes the model consider local detail information and spatial position information in the low-level features when dealing with high-level features, thereby alleviating the interference of background information.

    In ViT\cite{dosovitskiy2020image}, the patch merge layer is generally used to implement downsampling, which effectively increases the channel extension of the layer to avoid information loss. 
    In addition, Convnext\cite{liu2022convnet} discussed in detail how to use ViT design to adapt to CNN. Inspired by patch merge and attention, we propose local spatial information enhancement module LSE.
    LSE uses the patch merge method to extract low-level information, and then integrates the extracted information into high-level features. 
    This method not only effectively fuses information of different scales, but also retains the original features of each level and avoids the introduction of noise. 
    As shown in Fig. \ref{fig:lse}, LSE is divided into two branches. The first
    branch is the low-level feature weight extraction operation, and the second branch
    is to fuse the extracted low-level feature weight into high-level features,
    so that high-level features pay more attention to object information and
    reduce background interference.

    Specifically, the input of LSE comes from two parts, one is the high-level feature $x_{H}$, and the other is the low-level feature $x_{L}$. 
    Both are multiscale features extracted from the backbone. They are used as inputs to the LSE to enhance the low-level information into the high-level features, and finally output the enhanced high-level features. 
    Firstly, the $3 \times 3$ convolution operation is performed on the high-resolution features of the low-level features, and then the patch merge operation is performed on the features obtained by the convolution. 
    The role of patch merge is to achieve downsampling while effectively increasing channels to avoid loss of local information. Depending on the size of the high-level feature $H^{\prime} \times W^{\prime}$, the block operation is performed to obtain four patches, and then the concatenate operation is performed on the four patches.
    Finally, the feature map of $4 \times C^{\prime} \times H^{\prime} \times W^{\prime}$ is obtained, and then the point convolution is performed on the features to realize the information interaction between different channels, making its output channel number consistent with the number of channels that need to be enhanced at the high level. The formula is as follows:

    \begin{equation}
         x_{w}=Conv_{1\times1}(PatchMerge(Conv_{3\times3}(x_{L}))).
    \end{equation}
       
    The secondary branch leverages the low-level local spatial information weights generated by the primary branch to perform selective fusion with high-level channel features, ultimately producing enhanced high-level feature representations.
    First, the high-level features are divided into channels to obtain the channels that need to be enhanced and the channels that remain unchanged. Then, the low-level
    feature weights are element-wise multiplied by the high-level features that need to be enhanced.
    Finally, the two parts are concatenated to obtain enhanced high-level features.
    The formula is as follows:

    \begin{equation}
    \begin{aligned}
        &x^{fuse}_{H},x^{identity}_{H}=split(x_{H}), \\
        &concatenate(x_{w}\cdot x^{fuse}_{H},x^{identity}_{H}).
    \end{aligned}
    \end{equation}

    \begin{figure}[t]
        \centering
        \includegraphics[width=0.8\textwidth]{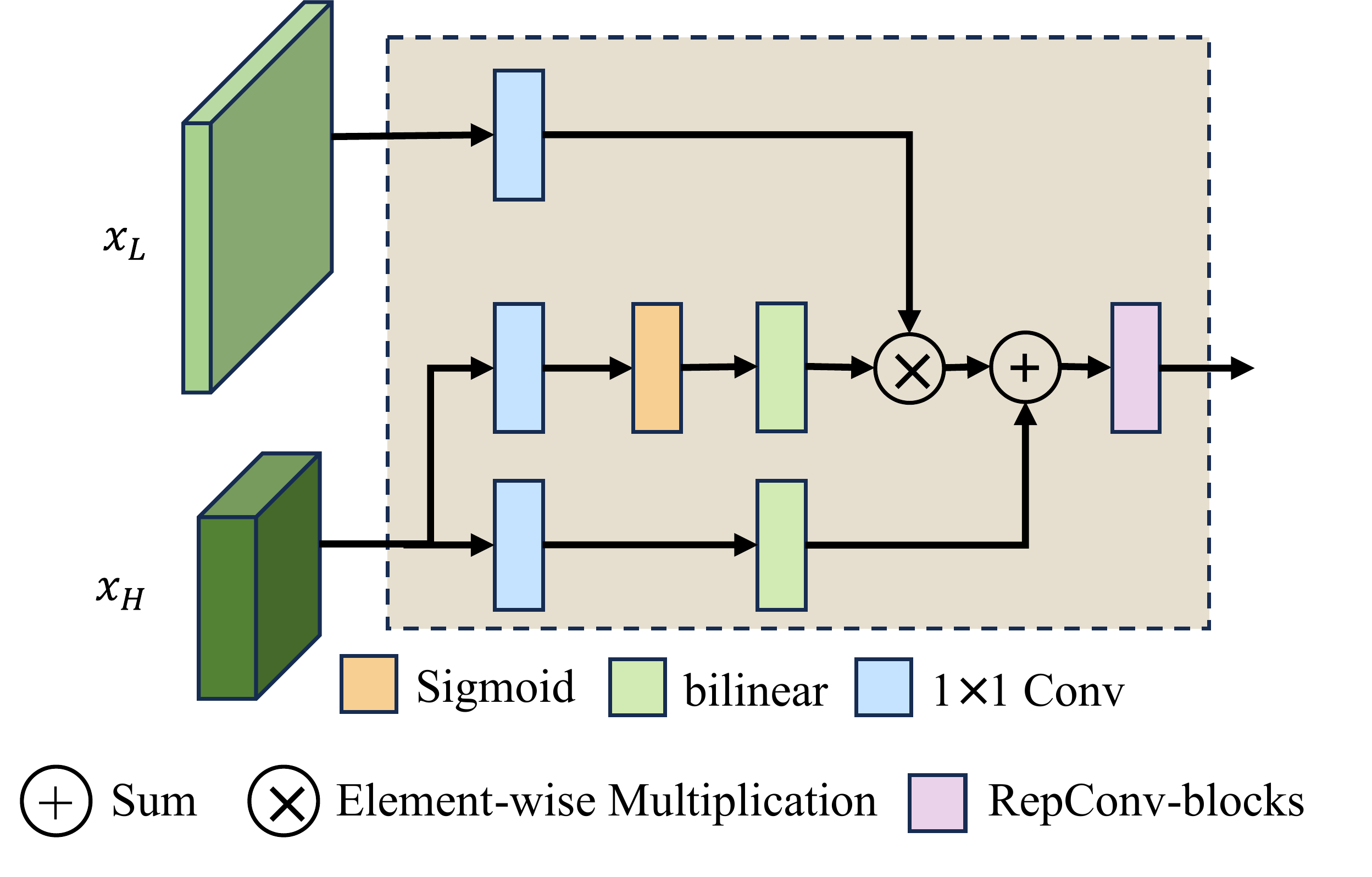}
        \caption{Overview of GII, which comprises two branches. The first branch is used to obtain information about local spatial weights and global semantics, and
        the second branch is to fuse the extracted information into
        low-level features.}
        \label{fig:gii}
    \end{figure}

    \subsection{Global Information Injection} 
    \label{met:GII} In the multiscale feature fusion part, the multiscale features output by the encoder are concatenated by the channel dimension. 
    For example, the feature set $( F_{L}, F_{M}, F_{H})$ output by the encoder, as shown in Fig.~\ref{fig:lgi-detr}, where $F_{L}, F_{M},$ and $F_{H}$ represent the feature maps from low, middle, and high layers, respectively. 
    The formula of multi-scale information series of RT-DETR is as follows:
    \begin{equation}
        F_{fuse}=Concatenate(F_{L},F_{M},F_{H}).
    \end{equation}
    In the decoder part, the following operations are performed on the multi-scale
    feature map generated by the encoder. Firstly, high-quality initial queries are
    generated by dynamic selection, and then the self-attention mechanism is
    performed on the object queries generated by dynamic selection to process
    the relationship between query objects. 
    Then the cross-attention mechanism is performed to correlate the multi-scale image feature information output by the query and the encoder part. 
    Finally, the features through the cross-attention mechanism are input to the detection head to locate and classify the prediction box. 

    The characteristic quality of the encoder output plays a vital role in the detection of the decoder prediction. 
    The low-level high-resolution feature map includes more local spatial information, and the high-level low-resolution feature map includes more global semantic information. 
    With the increase of convolution operations, high-level features are more likely to lose local related representation information, while low-level features intend to lack global semantic information.
    The lack of these important information inevitably leads to the low quality of the feature maps generated by the model.
    The limitation of this method is particularly obvious in small object detection tasks. 
    Due to the relatively small proportion of small objects in the image, their local information plays a vital role in improving the detection accuracy. 
    However, only using low-level features for prediction often fails to correlate global semantic information well. 
    Therefore, the performance of the model will decrease when detecting small objects.

    To address these limitations, we propose a Global Information Injection (GII) module, specifically designed to compensate for the deficiency of global semantic information in low-level features.
    At the final stage of the efficient encoder, the proposed method enhances low-level features by injecting global semantic information derived from high-level features. 
    In order to more effectively inject high-level information into low-level feature information, we are inspired by \cite{wan2023seaformer} and use attention operations for feature fusion as shown in the Fig.~\ref{fig:gii}. 
    This model that combines low-level features with high-level features can better adapt to multi-scale information and complex scenes, thereby significantly improving the flexibility and adaptability of the model.

    Specifically, the input of GII contains two parts, one is the low-level
    feature $x_{L}$, and the other is the high-level feature $x_{H}$. Firstly, two
    different $1 \times 1$ convolutions are used for $x_{H}$. One branch obtains
    high-level semantic weight information $W_{H}$ with the same low-level
    resolution through sigmoid activation and bilinear interpolation, while the other
    branch only obtains high-level semantic information $I_{H}$ with the same low-level
    resolution through bilinear interpolation. Then, a $1 \times 1$ convolution is
    applied to the low-level information, multiplied by the high-level semantic
    weight information $W_{H}$, and added to the high-level global semantic information
    $I_{H}$ to obtain the fused low-level features $x_{att\_fuse}$ with high-level
    global information. Finally, the feature information is further extracted and
    fused by RepBlock. The structure of the GII can be formally expressed as follows:

    \begin{equation}
    \begin{aligned}
        &W_{H}=resize(Sigmoid(Conv_{1\times1}(x_{H}))), \\
        &I_{H}=reisze(Conv_{1\times1}(x_{H})), \\
        &x_{att\_fuse}=Conv_{1\times1}(x_{L})\cdot W_{H}+ I_{H}, \\
        &x_{fuse}=RepBlock(x_{att_fuse}).
    \end{aligned}
    \end{equation}
    where $resize(\cdot)$ represent bilinear interpolation.

\section{Experiment and Analysis}
    \label{sec:exp} 
    To validate the superiority of the proposed LGI-DETR, it is compared with multiple
    state-of-the-art approaches on two object detection datasets, namely, VisDrone
    and UAVDT.

    \subsection{Datasets}
    \textbf{\emph{VisDrone-2019-DET}} The VisDrone-2019-DET dataset comprises 6,471
    training images, 548 validation images, and 3,190 test images, all captured
    from drones at varying altitudes at different locations. Each image is annotated
    with bounding boxes for ten predefined object categories: pedestrian, person,
    car, van, bus, truck, motorbike, bicycle, awning-tricycle, and tricycle. We
    used the VisDrone-2019-DET training set and validation set for training and
    testing, respectively.

    \textbf{\emph{UAVDT}} The UAVDT dataset comprises 23258 training images and
    15069 testing images. The UAVDT dataset contains 38327 UAV aerial images (23258
    training images and 15069 testing images). Its scenes are similar to those
    in VisDrone2019, encompassing three categories: car, bus, and truck. We choose
    the training set for training and the test set for testing.

    \subsection{Evaluation Metrics} 
    The standard COCO metrics we use to evaluate and compare the performance of various methods, including the AP(averaged over uniformly sampled IoU thresholds ranging from 0.50-0.95 with a step size of 0.05), and AP$_{50}$ (AP at an IoU threshold of 0.50). 
    Additionally, to comprehensively evaluate the model, metrics such as GFLOPs, and parameters are employed to determine the model's complexity. 
    The GFLOPs are calculated based on an input resolution of 640 $\times$ 640.

    \subsection{Implementation Details}
    Our model is developed based on the RT-DETR codebase. 
    Model is trained from scratch on NVIDIA RTX 3090 GPU. The training of LGI-detr is executed using the AdamW optimizer with learning rate 0.0001, weight decay of 0.0001.  We train the model for 100 epochs with a batch size of 8. We use an early stopping mechanism with patience of 20. The input image size is scaled to 640$\times$640 pixels. Additionally,We adopted a phased data enhancement method. 
    In the last 25 epochs, we closed the mosaic data enhancement and only retained the translate, fliplr, scale and hsv data augmentation methods.

    \begin{table}[t]
        \caption{Experimental Results on the VisDrone2019-val Dataset.}
        \begin{center}
        \begin{tabular}{l c c c c c} 
            \hline
            \textbf{Model}                    & \textbf{Publication} & \textbf{Params(M)} & \textbf{GFLOPs} & \textbf{AP}   & \textbf{AP$_{50}$} \\
            \hline
            YOLOv8-M\cite{yolov8_ultralytics} & -                    & 25.9               & 78.9            & 24.6          & 40.7               \\
            YOLOv9-M\cite{wang2024yolov9}     & ECCV2024             & 20.1               & 76.8            & 25.2          & 42.0               \\
            YOLOv10-M\cite{wang2024yolov10}   & NeuralIPS2024        & 15.4               & 59.1            & 24.5          & 40.5               \\
            Gold-YOLO-S\cite{wang2023gold}    & NeuralIPS2023        & 21.5               & 46.0            & 23.3          & 38.6               \\
            HIC-YOLOv5\cite{tang2024hic}      & ICRA2024             & 9.4                & 31.2            & 20.8          & 36.1               \\
            EdgeYOLO-S\cite{liu2023edgeyolo}  & CCC2023              & 9.3                & 41.0            & 23.6          & 40.8               \\
            RT-DETR-r18\cite{zhao2024detrs}   & CVPR2024             & 20.0               & 60.0            & 26.2          & 43.6               \\
            \hline
            \textbf{LGI-DETR(Ours)}           & -                    & 21.1               & 65.0            & \textbf{28.1} & \textbf{46.0}      \\
            \hline
        \end{tabular}
        \end{center}
        \label{table:visdrone}
    \end{table}

    \subsection{Comparison with State-of-the-art Methods}

    \subsubsection{Comparisons on VisDrone2019-DET}
    Our proposed model demonstrates significant improvements over existing models
    in terms of the key evaluation metric, mean Average Precision (mAP), on the VisDrone
    dataset. Table \ref{table:visdrone} illustrate a comparative analysis of the
    performance of our proposed method against other leading YOLOs, UAV detectors
    and baseline RT-DETR on the VisDrone2019-val dataset. Compared to baseline
    model RT-DETR-r18, our proposed method achieves 1.9\% , 2.4\% improvement in
    AP and AP$_{50}$, respectively. We also compare our proposed method with
    YOLO-serials detectors, such as YOLOv8\cite{yolov8_ultralytics}, YOLOv9\cite{wang2024yolov9}, YOLOv10\cite{wang2024yolov10} and Gold-YOLO\cite{wang2023gold}.
    Compared with YOLOv8-M, YOLOv9-M, YOLOv10-M and Gold-YOLO-S, LGI-DETR improves
    AP by 3.5\%, 2.9\%, 3.6\%, and 4.8\%, respectively. Furthermore, we compared
    our method with other UAV-object detectors, such as HIC-YOLOv5\cite{tang2024hic} and EdgeYOLO-S\cite{liu2023edgeyolo}, and the results show that our method also outperforms the others in terms of accuracy.

    \begin{table}[tbp]
        \caption{Experimental Results on the UAVDT-test Dataset.}
        \centering
        \footnotesize
        \begin{tabular}{l c c c c c c} 
            \hline
            \textbf{Model}          & \textbf{Params(M)} & \textbf{GFLOPs} & \textbf{AP}   & \textbf{AP$_{50}$} \\
            \hline
            Gold-YOLO-S\cite{wang2023gold}             & 21.5               & 46.0            & 14.9          & 26.5               \\
            EdgeYOLO-S\cite{liu2023edgeyolo}              & 9.3                & 41.0            & 15.9          & 29.4               \\
            RT-DETR-r18\cite{zhao2024detrs}             & 20.0               & 60.0            & 16.3          & 29.1               \\
            \hline
            \textbf{LGI-DETR(Ours)} & 21.1               & 65.0            & \textbf{17.4} & \textbf{30.2}      \\
            \hline
        \end{tabular}%
        \label{table:uavdt}
    \end{table}

    \subsubsection{Comparisons on UAVDT}
    As depicted in Table \ref{table:uavdt}, we also compare LGI-DETR's performance
    with YOLOs\cite{liu2023edgeyolo,wang2023gold}, and baseline\cite{zhao2024detrs} method on the UAVDT dataset to evaluate the generalization ability of LGI-DETR. 
    The result show that our model reaches 17.4\% and 30.2\% on AP and AP$_{50}$, respectively. 
    This result exceeds other YOLO detectors. Compared with the baseline model,
    the AP and AP$_{50}$ were increased by 0.9\% and 1.1\%, respectively. This shows
    that the two feature information enhancement modules we proposed are very important
    for the accuracy of small object detection. This can effectively improve the
    model 's ability to locate and classify small objects in UAV images.

    \subsection{Ablation Studies and Analysis} 
    We designed a series of ablation experiments to verify the effectiveness of
    our proposed method. We evaluate our model from two aspects : detection accuracy
    and model complexity. The table \ref{table:abl} shows the performance
    comparison under different configurations, where LSE represents the Local Spatial
    Enhancement module and GII represents the Global Information Injection module.
    Baseline achieves 26.2\% AP and 43.6\% AP$_{50}$ on the VisDrone2019 dataset.
    The model's parameters and GFLOPs are 20.0M, 60. Next, we gradually add the two
    modules we proposed to the encoder part of the baseline. Firstly, we add our
    proposed LSE, which achieves 27.8\% AP and 45.4\% AP$_{50}$ on the accuracy metric.
    Compared with the baseline model, the model with the LSE
    module increased by 1.6\% and 1.8\% on AP and AP50, respectively. And The
    number of parameters and GFLOPs only increased by 0.4M and 2. It shows that our
    proposed LSE module can effectively enhance the low-level local spatial information
    into the high-level features. After that, we add the GII module on the baseline,
    and reached 26.7\%AP and 44.3\% AP$_{50}$ on the accuracy
    metric. And The number of parameters and GFLOPs only increased by 0.3M and
    2. It shows that our proposed global information injection module provides
    rich global semantic information for low-level features, so as to facilitate
    the object detection of the subsequent decoder head. Finally, the two modules
    proposed by us are integrated on the baseline, and finally the accuracy
    indicators reach 28.1\% AP and 46.0\% AP$_{50}$, which is
    improved 1.9\% and 2.4\% compared with the baseline model. And the number of
    parameters and GFLOPs are only increased by 1M and 5, respectively.

    \begin{table}[tbp]
        \caption{Results of the Ablation Study on VisDrone2019-val Dataset}
        \centering
        \footnotesize
        \begin{tabular}{c c c c c c c} 
            \hline
            \textbf{Baseline} & \textbf{LSE} & \textbf{GII} & \textbf{Params(M)} & \textbf{GFLOPs} & \textbf{AP}   & \textbf{AP$_{50}$} \\
            \hline
            \Checkmark        & \XSolidBrush & \XSolidBrush & 20.0               & 60              & 26.2          & 43.6               \\
            \Checkmark        & \Checkmark   & \XSolidBrush & 20.5               & 62              & 27.8          & 45.4               \\
            \Checkmark        & \XSolidBrush & \Checkmark   & 20.5               & 62              & 26.7          & 44.3               \\
            \hline
            \Checkmark        & \Checkmark   & \Checkmark   & 21.1               & 65              & \textbf{28.1} & \textbf{46.0}      \\
            \hline
        \end{tabular}
        \label{table:abl}
    \end{table}

    \begin{figure}[hb]
        \centering
        \includegraphics[width=\textwidth]{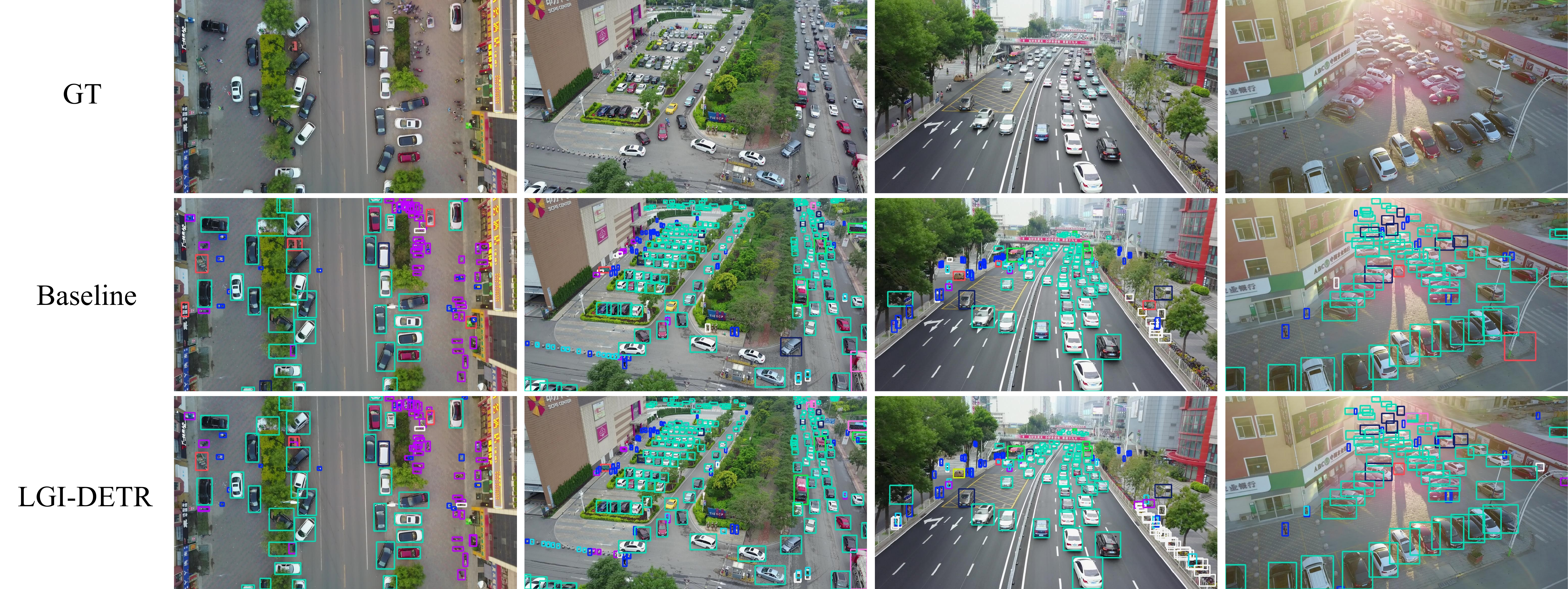}
        \caption{Visualization. In different scenarios, the detection performance
        of LGI-DETR and baseline is compared. First row is ground truth. Second row
        is baseline model's detection result. Third row is LGI-DETR's detection
        result.}
        \label{fig:vis}
    \end{figure}

    \subsection{Visualization}
    In Fig.~\ref{fig:vis}, we visualize the detection results of the baseline model
    and our proposed model. object detection in sparse, dense and high-exposure scenes.
    Compared with the baseline model, LGI-DETR has a significant improvement in small
    object detection, and has fewer false detections and missed detections. In addition,
    it can adapt to various complex detection situations. This fully demonstrates
    that LGI-DETR is more robust in object location and classification. This
    also shows that our proposed bidirectional cross-layer local global
    information enhancement module effectively improves the performance of UAV
    image object detection.

    \section{Conclusion}
    \label{sec:con} 
    We design a novel LGI-DETR model to improve the detection performance of UAV
    images. The model is based on RT-DETR, which is an efficient transformer-based
    end-to-end detector. We have carefully designed the encoder part. The model includes
    two key modules: the Local Spatial Enhancement module (LSE) and the Global
    Information Injection module (GII). Experimental results on the VisDrone demonstrate
    that our proposed method does not significantly increase the complexity of the
    model in UAV object detection, while the accuracy is greatly improved. However,
    in practical applications, our model still has some limitations. For example,
    our model still faces certain difficulties in detecting small objects in complex
    backgrounds. In the future, we will continue to explore ways to further improve
    the performance of the model to adapt to more scenarios.

%
%
\bibliographystyle{splncs04}
\bibliography{main.bib}

\end{document}